\documentclass[final]{cvpr}

\usepackage{times}
\usepackage{epsfig}
\usepackage{graphicx}
\usepackage{amsmath}
\usepackage{amssymb}

\usepackage{kotex}
\usepackage{url}
\usepackage{tabularx}
\usepackage{gensymb}
\usepackage{multirow}
\usepackage{tablefootnote}
\usepackage{makecell}

\usepackage[pagebackref=true,breaklinks=true,colorlinks,bookmarks=false]{hyperref}

\def\cvprPaperID{****} 
\def\confYear{CVPR 2021}
\begin{document}

\title{Perceptual Image Quality Assessment with Transformers}

\author{Manri Cheon, Sung-Jun Yoon, Byungyeon Kang, and Junwoo Lee\\
LG Electronics\\
{\tt\small \{manri.cheon, sungjun.yoon, byungyeon.kang, junwoo.lee\}@lge.com}
}

\maketitle
\thispagestyle{empty}
\begin{abstract}
In this paper, we propose an image quality transformer (IQT) that successfully applies a transformer architecture to a perceptual full-reference image quality assessment (IQA) task.
Perceptual representation becomes more important in image quality assessment.
In this context, we extract the perceptual feature representations from each of input images using a convolutional neural network (CNN) backbone.
The extracted feature maps are fed into the transformer encoder and decoder in order to compare a reference and distorted images.
Following an approach of the transformer-based vision models \cite{dosovitskiy2020image, you2020transformer}, we use extra learnable quality embedding and position embedding.
The output of the transformer is passed to a prediction head in order to predict a final quality score.
The experimental results show that our proposed model has an outstanding performance for the standard IQA datasets.
For a large-scale IQA dataset containing output images of generative model, our model also shows the promising results.
The proposed IQT was ranked first among 13 participants in the NTIRE 2021 perceptual image quality assessment challenge \cite{gu2021ntire}.
Our work will be an opportunity to further expand the approach for the perceptual IQA task.
\end{abstract}


\thispagestyle{empty}
\section{Introduction}
\label{sec1}

Perceptual image quality assessment (IQA) is an important topic in the multimedia systems and computer vision tasks \cite{chikkerur2011objective, sheikh2006statistical, zhai2020perceptual}.
One of the goals of the image processing is to improve the quality of the content to an acceptable level for the human viewers.
In this context, the first step toward generating acceptable contents is to accurately measure the perceptual quality of the content, which can be performed via subjective and objective quality assessment \cite{ wang2011applications, cheon2017subjective, hu2020subjective, fang2020perceptual}.
The subjective quality assessment is the most accurate method to measure the perceived quality, which is usually represented by mean opinion scores (MOS) from collected subjective ratings.
However, it is time-consuming and expensive.
Thus, objective quality assessment performed by objective metrics is widely used to automatically predict perceived quality \cite{wang2004image, wang2003multiscale, sheikh2006image, zhang2011fsim, xue2014gradient}.

However, with the recent advances in deep learning-based image restoration algorithms, accurate prediction of the perceived quality has become more difficult.
In particular, image restoration models based on generative adversarial network (GAN) \cite{gan} have been developed in order to improve the perceptual aspect of the result images \cite{wang2018esrgan, blau20182018, choi2020deep, cheon2018generative}.
However, it sometimes generates output images with unrealistic artifacts.
The existing objective metrics such as Peak Signal-to-Noise Ratio (PSNR), a structural similarity index (SSIM) \cite{wang2004image}, and conventional quality metrics are insufficient to predict the quality of this kind of outputs.
In this respect, recent works \cite{zhang2018unreasonable, ding2020image, gu2020image, prashnani2018pieapp, bosse2017deep} based on perceptual representation exhibit a better performance at the perceptual IQA task.
As various image restoration algorithms are developed, however, it is still required to develop the IQA algorithm that accurately predicts the perceptual quality of images generated by emerging algorithms.


\begin{figure*}[ht]
    \centering
	\includegraphics[width=1\textwidth]{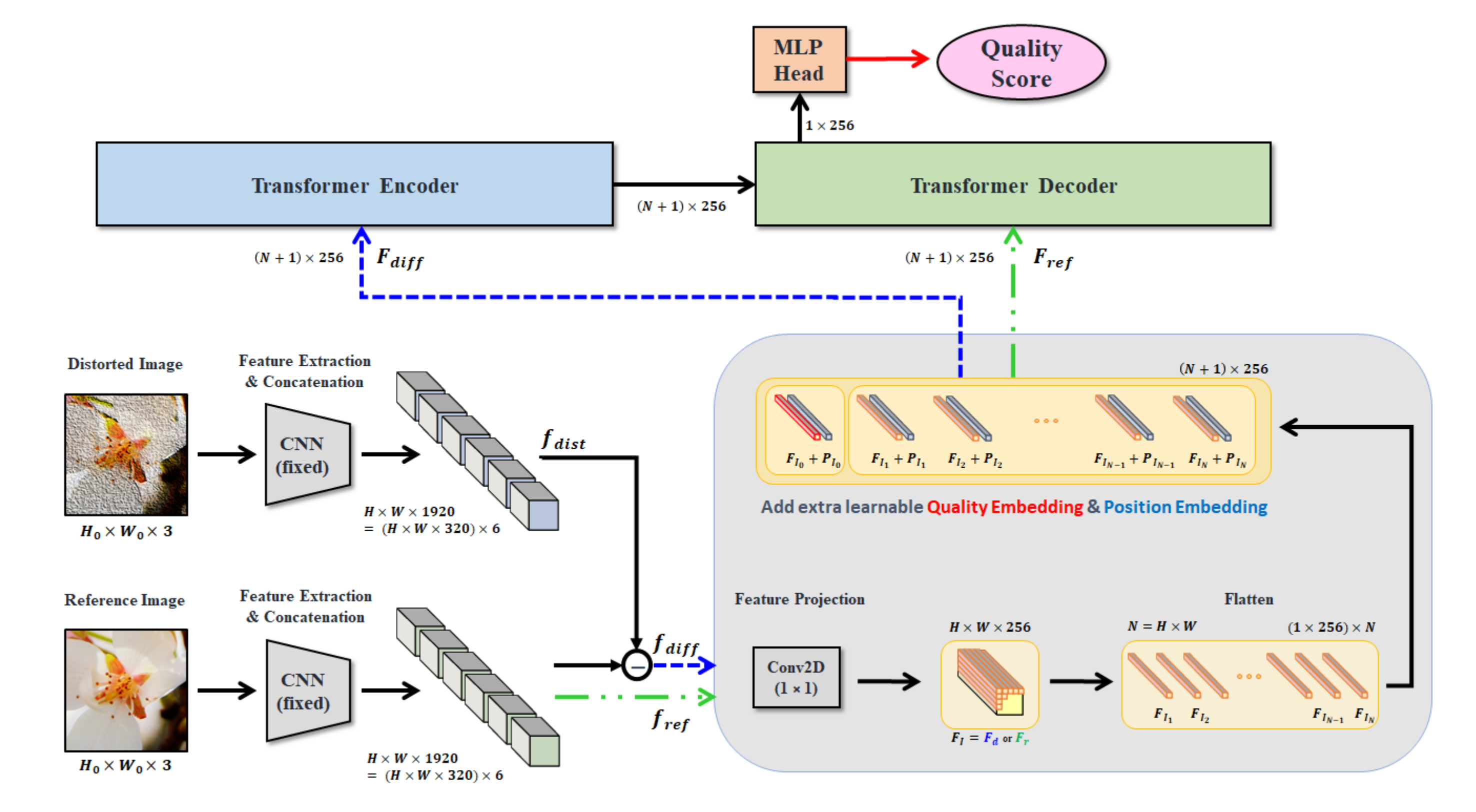}
	\caption{Model architecture of proposed image quality transformer (IQT). Note that $F_{I}$ denotes $F_{d}$ and $F_{r}$ in Eqs. \ref{eq:enc} and \ref{eq:dec}, respectively.}
	\label{fig:model}
\end{figure*}


In recent years, based on the success in the natural language processing (NLP) field, the transformer \cite{vaswani2017attention} architecture has been applied in the computer vision field \cite{khan2021transformers}.
A wider research area in the computer vision has been improved based on the transformer, such as recognition task \cite{carion2020end, touvron2020training, dosovitskiy2020image}, generative modelling \cite{parmar2018image, jiang2021transgan, chen2020generative}, low-level vision \cite{chen2020pre, yang2020learning, kumar2021colorization}, etc.
However, few attempts were made in the field of the image and video quality assessment.
In a recent study, You and Korhonen proposed the application of transformer in image quality assessment \cite{you2020transformer}.
They achieved outstanding performance on two publicly available large-scale blind image quality databases.
With our knowledge, however, this study is the only transformer-based approach for image quality assessment.
Therefore, it is urgently needed to investigate whether the transformer-based approach works well in the field of perceptual image quality assessment.
Especially, it should be investigated whether this structure is applicable to a full-reference (FR) model aiming to measure the perceptual similarity between two images.
In addition, it is also necessary to evaluate whether this approach can accurately predict the perceptual quality for the latest GAN-based artifacts.

In this study, we propose an Image Quality Transformer (IQT), which is the FR image quality assessment method as shown in Fig. \ref{fig:model}.
To tackle the perceptual aspects, a convolutional neural network (CNN) backbone is used to extract perceptual representations from an input image.
Based on the transformer encoder-decoder architecture, the proposed model is trained to predict the perceptual quality accurately.
The proposed model was ranked in the first place among 13 participants in the NTIRE 2021 challenge on perceptual image quality assessment \cite{gu2021ntire} at the CVPR 2021.

The rest of this article is organized as follows.
The following section presents the related work.
Section \ref{sec3} describes the proposed method and the experiments are given in Section \ref{sec4}.
Finally, conclusions are given in Section \ref{sec5}.

\thispagestyle{empty}
\section{Related Work}
\label{sec2}

\paragraph{Image Quality Assessment.}
The most important goal of the developing objective IQA is to accurately predict the perceived quality by human viewers.
In general, the objective IQA methods can be classified into three categories according to the existence of reference information: FR \cite{wang2004image, wang2003multiscale, sheikh2006image,larson2010most, chandler2007vsnr}, reduced-reference (RR) \cite{soundararajan2011rred}, and no-reference (NR) \cite{mittal2012making, ma2017learning} IQA methods.
The NR method is useful for the system because of its feasibility.
However, the absence of a reference makes it challenging to predict image quality accurately compared to the FR method.
The FR method focuses more on visual similarity or dissimilarity between two images, and this method still plays an important role in the development of image processing system.

The representative of the commonly and widely used quality FR metric is the PSNR.
It has the advantage of convenience for optimization; however, it tends to poorly predict perceived visual quality.
Wang \textit{et al.} proposed the SSIM \cite{wang2004image} that is based on the fact that the human visual system (HVS) is highly correlated to structural information.
Since that, various FR metrics have been developed to take into account various aspects of human quality perception, e.g., information-theoretic criterion \cite{sheikh2006image, sheikh2005information}, structural similarity \cite{wang2003multiscale, zhang2011fsim}, etc.
Recently, CNN-based IQA methods as well as other low-level computer vision tasks have been actively studied \cite{zhang2018unreasonable, bosse2017deep, prashnani2018pieapp, ding2021comparison, hosu2020koniq}.
Zhang \textit{et al.} proposed a learned perceptual image patch similarity (LPIPS) metric \cite{zhang2018unreasonable} for FR-IQA.
The LPIPS showed that trained deep features that are optimized by the Euclidean distance between distorted and reference images are effective for IQA compared to the conventional IQA methods.
Ding \textit{et al.} proposed the metric that is robust to texture resampling and geometric transformation based on spatial averages of the feature maps \cite{ding2020image}.
Various IQA methods including aforementioned metrics are included in our experiments for performance comparison.

The primary criterion of performance measurement is the accuracy of the metrics.
Pearson linear correlation coefficient (PLCC) followed by the third-order polynomial nonlinear regression \cite{sheikh2006statistical} is usually used in order to evaluate the accuracy of the methods.
Spearman rank order correlation coefficient (SRCC) and the Kendall rank order correlation coefficient (KRCC) are used to estimate the monotonicity and consistency of the quality prediction.
Additional statistical method \cite{krasula2016accuracy} and an ambiguity based approach have also been proposed in \cite{cheon2021ambiguity}.
In our study, we select the SRCC, KRCC, and PLCC as performance evaluation metrics.

\paragraph{Vision Transformer.}

The transformer \cite{vaswani2017attention} consists of multi-head attentions (MHAs), multi-layer perceptrons (MLPs), layer normalizations (LNs) \cite{ba2016layer}, and residual connections.
Unlike the CNN, the transformer has a minimum inductive bias and can scale with the length of the input sequence without limiting factors.
Recently, it has emerged that the transformer has combined with the CNN using the self-attention \cite{carion2020end}, and some of which have completely replaced CNN \cite{wang2020axial}.

The transformer is mainly self-attention based approach.
Since the self-attention layer aggregates global information from the entire input sequence, therefore, the model can capture the entire image for measuring the perceptual quality of the whole image.
Vision Transformer (ViT) \cite{dosovitskiy2020image} is a representative success model among transformer-based vision models.
A hybrid architecture was proposed for image recognition using a concord of CNN and the transformer encoder.
It replaces the pixel patch embedding with the patches extracted from the CNN feature map.
This architecture could be applied well in the IQA task, because the effectiveness of the deep features on the perceptual IQA task was demonstrated in recent studies \cite{zhang2018unreasonable, ding2020image, you2020transformer}.
In DETR \cite{carion2020end}, the encoder-decoder architecture is employed and the decoder takes learned positional embeddings as object queries for object detection.
This approach could be applied to the FR-IQA model that compares two images and measures the similarity.
To measure similarity, one of the two images can be adopted as the query information in the self-attention layer.
From the successful approaches using the transformer, we learn the direction to develop the perceptual IQA method with the transformer.

\paragraph{Vision Transformer based IQA.} 
Inspired by ViT, TRIQ \cite{you2020transformer} naturally attempts to solve the blind IQA task using the transformer with the MLP head.
In order to exploit ViT and handle images with different resolution, the TRIQ model defines the positional embedding with sufficient length to cover the maximal image resolution.
The transformer encoder employs adaptive positional embedding, which handles a image with arbitrary resolutions.
The output of the encoder is fed into the MLP head and the MLP head predicts the perceived image quality.

\begin{figure}[ht]
    \centering
	\includegraphics[width=0.9\linewidth]{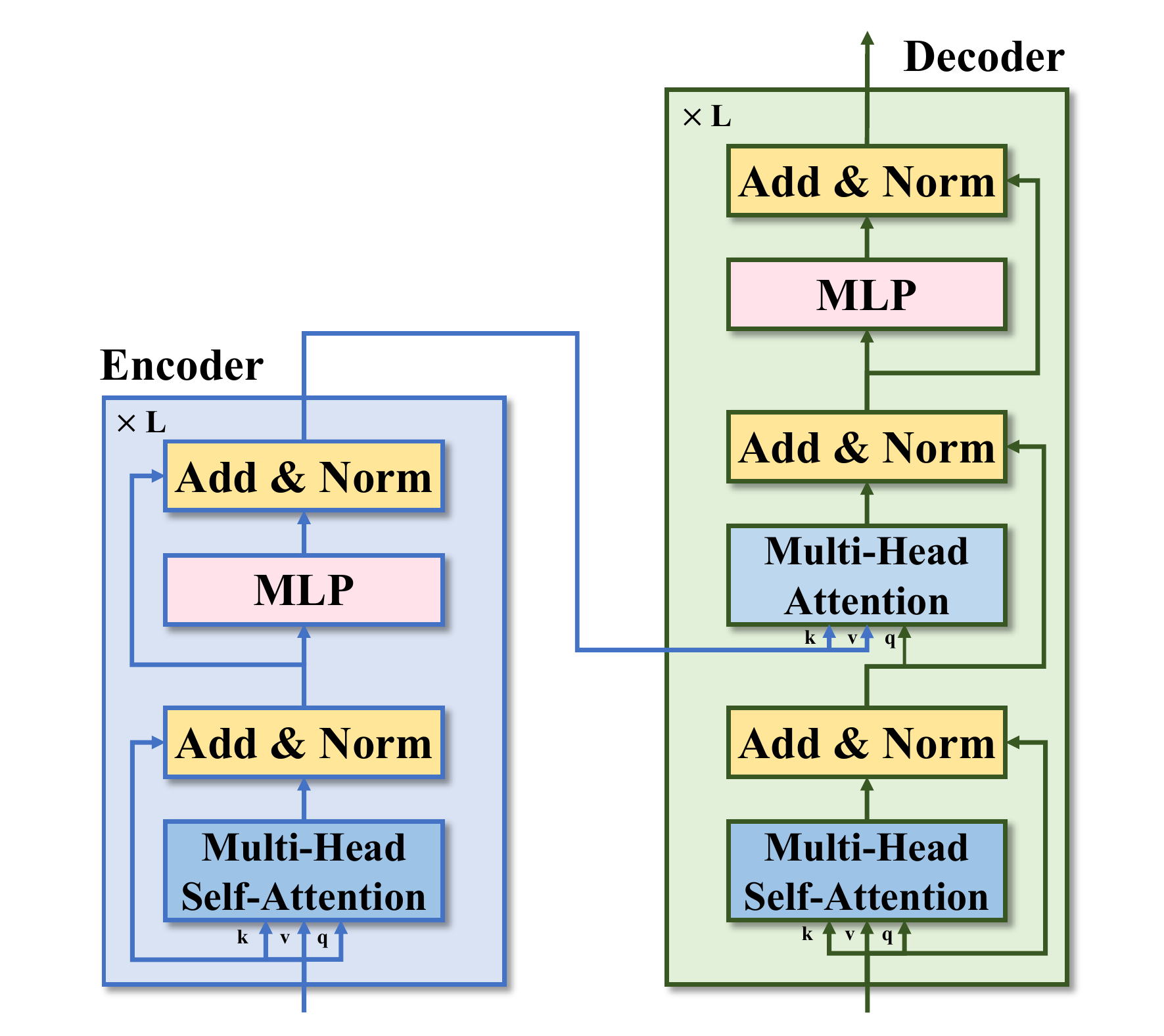}
	\caption{The transformer encoder and decoder.}
	\label{fig:transformer}
\end{figure}

Basically, similar to the TRIQ, our proposed model applies the transformer architecture for the IQA task.
However, additional aspects are considered in order to design the perceptual FR-IQA with the transformer. 
First, the transformer encoder-decoder architecture is an important point in our approach.
The reference information and the difference information between the distorted and reference images are employed as an input into the transformer.
Second, we adopt the Siamese architecture to extract both the input feature representations from the reference and distorted images.
For each image, by concatenating multiple feature maps extracted from intermediate layers, we obtained sufficient information for the model. 


\thispagestyle{empty}
\section{Proposed Method}
\label{sec3}

The proposed method that is illustrated in Fig. \ref{fig:model} consists of three main components: a feature extraction backbone, a transformer encoder-decoder, and a prediction head.
First, we use a CNN backbone to extract feature representations from both reference and distorted input images.
The extracted feature maps are projected to fixed size of vectors and flattened.
In order to predict perceived quality, the trainable extra \texttt{[quality]} embedding is added to the sequence of embedded feature.
It is similar to approach using \texttt{[class]} token in previous transformer models \cite{devlin2018bert, dosovitskiy2020image, touvron2020training}.
The position embedding is also added in order to maintain the positional information.
We pass this input feature embedding into the transformer encoder and decoder.
The transformer encoder and decoder are based on the standard architecture of the transformer \cite{vaswani2017attention}, where the structure is briefly illustrated in Fig. \ref{fig:transformer}.
The first vector of the output embedding of the decoder is fed into the MLP head in order to predict a single perceptual quality score.

\begingroup
\renewcommand{\arraystretch}{.88}
\begin{table*}[ht]
    \small
	\centering
	\caption{\label{tb:dataset} IQA datasets for performance evaluation and model training.}
	\begin{tabularx}{0.9\textwidth}{@{\extracolsep{\fill}}ccccccccc}
		\hline
		Database    & \# Ref.  &  \# Dist.   &   Dist. Type    &   \# Dist. Type &  \# Rating &   Rating Type    &   Env.\\ \hline
		LIVE \cite{sheikh2006statistical}	&	29	    &	779     & traditional        & 5   & 25k	& MOS &	lab\\
		CSIQ \cite{larson2010most}	&	30	    &	866	    & traditional        & 6   & 5k	    & MOS &	lab\\
		TID2013 \cite{ponomarenko2015image}	&	25	    &	3,000	& traditional        & 25  & 524k	& MOS &	lab\\ 
		KADID-10k \cite{kadid10k}	&	81	    &	10.1k	& traditional & 25  & 30.4k	& MOS &	crowdsourcing\\
		PIPAL \cite{pipal}	&	250	    &	29k	    & trad.+alg. outputs & 40  & 1.13m	& MOS &	crowdsourcing\\ \hline
	\end{tabularx}
\end{table*}
\endgroup

\paragraph{Feature Extraction Backbone.}
A conventional CNN network, Inception-Resnet-V2 \cite{szegedy2017inception}, is employed as the feature extraction backbone network.
Pretrained weights on ImageNet \cite{deng2009imagenet} is imported and frozen.
Feature maps from six intermediate layers of Inception-Resnet-V2, i.e., \{$mixed\_5b$, $block35\_2$, $block35\_4$, $block35\_6$, $block35\_8$, $block35\_10$\}, are extracted.
The extracted feature maps have the same shape $f_{layer} \in \mathbb{R}^{H \times W \times c}$, where $c=320$, and they are concatenated into feature map.
In other words, for an input image  $I \in \mathbb{R}^{H_{0} \times W_{0} \times 3}$, the feature map $f \in \mathbb{R}^{H \times W \times C}$, where $C = 6 \times c$, is extracted.

Both \textit{reference} and \textit{distorted} images are used; therefore, the two input feature maps, $f_{ref}$ and $f_{dist}$, are employed for the transformer, respectively.
In order to obtain difference information between reference and distorted images, a \textit{difference} feature map, $f_{d}$, is also used.
It can be simply obtained by subtraction between two feature maps of reference and distorted images, i.e., $f_{diff}=f_{ref}-f_{dist}$.

\paragraph{Transformer encoder.}
A \textit{difference} feature embedding, $F_{d} \in \mathbb{R}^{N \times D}$, is used as the input of the transformer encoder.
We first reduce the channel dimension of the $f_{d}$ to the transformer dimension $D$ using a $ 1 \times 1 $ convolution.
Then, we flatten the spatial dimensions, which means the number of patches in the feature map is set as $N = H\times W$.
As often used in the vision transformer models \cite{dosovitskiy2020image, you2020transformer}, we append extra quality embedding at the beginning of the input feature embedding as $F_{d_{0}}$.
And the trainable position embedding $P_{d} \in \mathbb{R}^{(1+N) \times D}$ are also added in order to retain the positional information.
The calculation of the encoder can be formulated as 
\begin{equation} \label{eq:enc}
\begin{split}
 y_{0} = [F_{d_{0}}+P_{d_{0}} , F_{d_{1}}+P_{d_{1}}, ... , F_{d_{N}}+P_{d_{N}}], \\
 q_{i} = k_{i} = v_{i} = y_{i-1}, \\
 y_{i}' = LN(MHA(q_{i},k_{i},v_{i})+y_{i-1}), \\
 y_{i} = LN(MLP(y_{i}')+y_{i}'),  \quad i = 1, ..., L \\
 [F_{E_{0}}, F_{E_{1}}, ... , F_{E_{N}}] = y_{L}, \\
\end{split}
\end{equation}
where $L$ denotes the number of the encoder layers.
The output of the encoder $F_{E} \in \mathbb{R}^{(1+N) \times D}$ has the same size to that of the input feature embedding.

\paragraph{Transformer decoder.}
The decoder takes the \textit{reference} feature embedding $F_{r} \in \mathbb{R}^{N \times D}$, obtained through the channel reduction and flattening.
The extra quality embedding and position embedding are also added to it.
The output of the encoder, $F_{E}$, is used as an input of the decoder, and it is used as a key-value in the second MHA layer.
The calculation of the decoder can be formulated as:
\begin{equation} \label{eq:dec}
\begin{split}
 y_{L} = [F_{E_{0}}, F_{E_{1}}, ... , F_{E_{N}}], \\
 z_{0} = [F_{r_{0}}+P_{r_{0}} , F_{r_{1}}+P_{r_{1}}, ... , F_{r_{N}}+P_{r_{N}}], \\
 q_{i} = k_{i} = v_{i} = z_{i-1}, \\
 z_{i}' = LN(MHA(q_{i},k_{i},v_{i})+z_{i-1}), \\
 q_{i}' = z_{i}', \quad k_{i}' = v_{i}' = y_{L}, \\
 z_{i}'' = LN(MHA(q_{i}',k_{i}',v_{i}')+z_{i}'), \\
 z_{i} = LN(MLP(z_{i}'')+z_{i}''), \quad i = 1, ..., L \\
 [F_{D_{0}}, F_{D_{1}}, ... , F_{D_{N}}] = z_{L},  \\
\end{split}
\end{equation}
where $L$ denotes the number of decoder layers.
The output embedding $F_{D} \in \mathbb{R}^{(1+N) \times D}$ of the decoder is finally obtained.

\paragraph{Head.}
The final quality prediction is computed in the prediction MLP head.
The first vector of the decoder output, $F_{D_{0}} \in \mathbb{R}^{1 \times D}$ in Eq. \ref{eq:dec}, is fed into the MLP head, which contains the quality information.
The MLP head consists of two fully connected (FC) layers, and the first FC layer is used followed by the ReLU activation.
The second FC layer has one channel to predict a single score.

\thispagestyle{empty}
\section{Experiments}
\label{sec4}

\begingroup
\renewcommand{\arraystretch}{.9}
\begin{table*}[ht]
	\centering
	\small
	\caption{\label{tb:bench1} Performance comparison of the IQA methods on three standard IQA databases, i.e., LIVE \cite{sheikh2006statistical}, CSIQ \cite{larson2010most}, and TID2013 \cite{ponomarenko2015image}, in terms of SRCC and KRCC. The top three performing methods are highlighted in bold face. Some results are borrowed from \cite{ding2020image, gu2020image}.}
	\begin{tabularx}{0.9\textwidth}{@{\extracolsep{\fill}}lcccccc}
		\hline
		\multirow{2}{*}{Method}&\multicolumn{2}{c}{LIVE\cite{sheikh2006statistical}}&\multicolumn{2}{c}{CSIQ\cite{larson2010most}}&\multicolumn{2}{c}{TID2013\cite{ponomarenko2015image}}\\ \cline{2-3} \cline{4-5} \cline{6-7}
		&SRCC&KRCC&SRCC&KRCC&SRCC&KRCC\\ \hline
		PSNR	&	0.873	&	0.680 	&	0.810 	&	0.601	&	0.687	&	0.496	\\
		SSIM \cite{wang2004image}	&	0.948	&	0.796	&	0.865	&	0.680	&	0.727	&	0.545	\\
		MS-SSIM \cite{wang2003multiscale}	&	0.951	&	0.805	&	0.906	&	0.730	&	0.786	&	0.605	\\
		VSI \cite{zhang2014vsi}	&	0.952	&	0.806	&	0.943	&	0.786	&	\textbf{0.897}	&	\textbf{0.718}	\\
		MAD \cite{larson2010most}	&	\textbf{0.967}	&	\textbf{0.842}	&	\textbf{0.947}	&	\textbf{0.797}	&	0.781	&	0.604	\\
		VIF \cite{sheikh2006image} &		0.964	&	0.828	&	0.911	&	0.743	&	0.677	&	0.518	\\
		FSIMc \cite{zhang2011fsim}	&	\textbf{0.965}	&	\textbf{0.836}	&	0.931	&	0.769	&	0.851	&	0.667	\\
		NLPD \cite{laparra2016perceptual}	&	0.937	&	0.778	&	0.932	&	0.769	&	0.800	&	0.625	\\
		GMSD \cite{xue2014gradient}	&	0.960	&	0.827	&	\textbf{0.950}	&	\textbf{0.804}	&	0.804	&	0.634	\\ \hline
		WaDIQaM \cite{bosse2017deep}	&	0.947	&	0.791	&	0.909	&	0.732	&	0.831	&	0.631	\\
		PieAPP \cite{prashnani2018pieapp}	&	0.919	&	0.750	&	0.892	&	0.715	&	\textbf{0.876}	&	\textbf{0.683}	\\
		LPIPS \cite{zhang2018unreasonable}	&	0.932	&	0.765	&	0.876	&	0.689	&	0.670	&	0.497	\\
		DISTS \cite{ding2021comparison}	&	0.954	&	0.811	&	0.929	&	0.767	&	0.830	&	0.639	\\
		SWD \cite{gu2020image}	&	-	&	-	&	-	&	-	&	0.819	&	0.634	\\ \hline
		IQT (ours)	& \textbf{0.970}	& \textbf{0.849} &  \textbf{0.943}    &   \textbf{0.799}   &  \textbf{0.899}   &   \textbf{0.717}   \\ 
		IQT-C (ours)	&   0.917	& 0.737 & 0.851 & 0.649 & 0.804 & 0.607 \\ \hline
	\end{tabularx}
\end{table*}
\endgroup

\thispagestyle{empty}
\subsection{Datasets}
We employ five databases that are commonly used in the research of perceptual image quality assessment.
The LIVE Image Quality Assessment Database (LIVE) \cite{sheikh2006statistical}, the Categorical Subjective Image Quality (CSIQ) database \cite{larson2010most}, and the TID2013 \cite{ponomarenko2015image} are the databases that serve as baselines for full-reference IQA studies.
These datasets only include traditional distortion types and the subjective scores are measured in the controlled laboratory environment.
KADID-10k \cite{kadid10k} is a large-scale IQA dataset and is chosen as the training dataset in our experiment.
It is three times larger compared to the TID2013 \cite{ponomarenko2015image} and the ratings are collected from crowdsourcing.
The PIPAL \cite{pipal} dataset is used for both the training and evaluation of the model in this study.
A large quantity of distorted images including GAN based algorithms' outputs and following human ratings are included in the PIPAL dataset.
It is challenging for existing metrics to predict perceptual quality accurately \cite{gu2020image}.
Table \ref{tb:dataset} summarizes the characteristics of the datasets employed in this study.

\thispagestyle{empty}
\subsection{Implementation details}
We denote our model trained on the KADID-10k as IQT.
The hyper-parameters for the model are set as follow: \romannumeral 1 ) the number of encoder and decoder layer is set to 2 (i.e., $L=2$), \romannumeral 2 ) the number of heads in the MHA is set to 4 (i.e., $H=4$), \romannumeral 3 ) the transformer dimension is set to 256 (i.e., $D=256$), \romannumeral 4 ) dimension of the MLP in the encoder and decoder is set to 1024 (i.e., $D_{feat}=1024$), \romannumeral 5 ) the dimension of the first FC layer in MLP head is set to 512 (i.e., $D_{head}=512$).

In the training phase, a given image is cropped to obtain image patches.
The dimension of the patch fed into the proposed IQT is $256\times256\times3$.
The number of patches in the feature map is set to $N=891$.
In the testing phase, image patches are also acquired from the given image pair.
We extract $M$ overlapping patches and predict final quality score by averaging $M$ individual quality scores of the patches.
The stride size is set as large as possible to cover the entire image with fewer patches.

Data augmentation including horizontal flip and random rotation is applied during the training.
The training is conducted using an ADAM \cite{kingma2015adam} optimizer with a batch size of 16.
Initial learning rate $2\times10^{-4}$ and cosine learning rate decay are set.
The training loss is computed using a mean squared error (MSE) loss function.
Our network is implemented using Tensorflow framework.
It roughly takes a half day with a single NVIDIA TITAN RTX to train our model.

\thispagestyle{empty}
\subsection{Results}

\begin{figure}[ht]
    \small
    \centering
	\includegraphics[width=.9\linewidth]{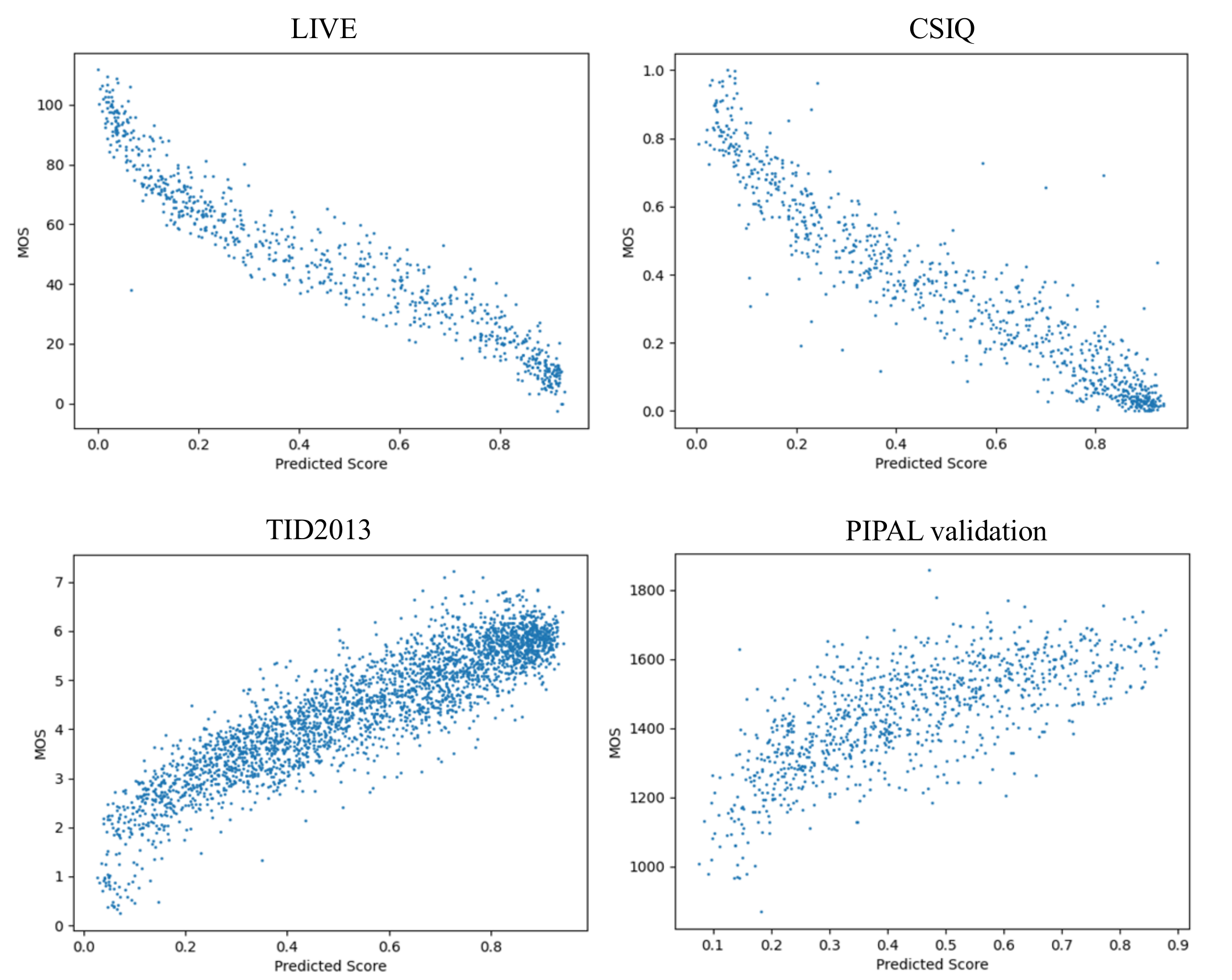}
	\caption{Scatter plots of ground-truth mean opinion scores (MOSs) against predicted scores of proposed IQT on LIVE, CSIQ, TID2013, and PIPAL datasets. The predicted scores are obtained from the model trained on KADID-10k dataset.}
	\label{fig:scatter1}
\end{figure}

\begin{figure*}[ht]
    \small
    \centering
	\includegraphics[width=1\linewidth]{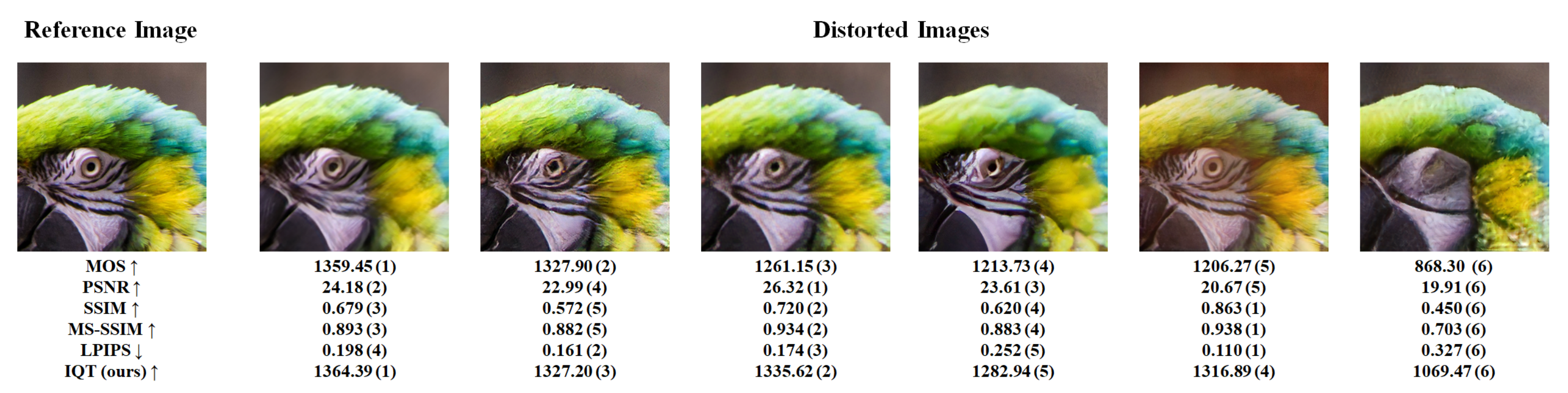}
	\caption{Example images from validation dataset of the NTIRE 2021 challenge. For each distorted image, predicted scores of PSNR, SSIM \cite{wang2004image}, MS-SSIM \cite{wang2003multiscale}, LPIPS \cite{zhang2018unreasonable}, and proposed IQT are listed. MOS denotes the ground-truth human rating. The number in the parenthesis denotes the rank among considered distorted images in this figure.}
	\label{fig:pipal1}
\end{figure*}

The proposed IQT shows that the transformer based model is sufficiently competitive compared to existing approaches for the dataset that has traditional distortions.
Our model is trained on KADID-10k and, then, the performance on the three standard IQA datasets is evaluated.
The performance comparison result is reported in Table \ref{tb:bench1} and the scatter plots of the predicted scores of IQT and the ground-truth MOS are also presented in Fig. \ref{fig:scatter1}.
For LIVE and TID2013 databases, the proposed IQT shows the best performance in terms of SRCC.
Also, it is ranked in the top three in all benchmarks in terms of SRCC and KRCC.
In particular, our method shows better performance than recent deep learning-based methods \cite{bosse2017deep, prashnani2018pieapp, zhang2018unreasonable, ding2020image, gu2020image} for most cases.

Example images of the PIPAL validation dataset and following PSNR, SSIM \cite{wang2004image}, MS-SSIM \cite{wang2003multiscale}, LPIPS \cite{zhang2018unreasonable}, and proposed IQT are illustrated in Fig. \ref{fig:pipal1}.
From the left to the right, the perceptually better images to worse images are listed based on MOS.
Our proposed IQT predicts the quality scores similar to MOS in terms of the superiority.
There exist the images that are clearly distinguished by all methods, however, it is difficult to accurately predict perceptual quality for some images.

\begingroup
\renewcommand{\arraystretch}{.9}
\begin{table}[b]
    \small
	\centering
	\caption{\label{tb:bench4} Performance comparison of IQA methods on PIPAL \cite{pipal} dataset. Main score is calculated with summation of PLCC and SRCC. The top performing method is highlighted in bold. Some results are provided from the NTIRE 2021 IQA challenge report \cite{gu2021ntire}.}
	\begin{tabularx}{.9\linewidth}{@{\extracolsep{\fill}}lcccc}
		\hline
		\multirow{2}{*}{Method} & \multicolumn{2}{c}{Validation} & \multicolumn{2}{c}{Testing} \\ \cline{2-3} \cline{4-5}
		 &    PLCC &   SRCC   & PLCC &   SRCC   \\ \hline
        PSNR	                            &   0.292   &   0.255   &   0.277	&	0.249	\\
        SSIM \cite{wang2004image}	        &   0.398   &   0.340   &   0.394	&	0.361	\\
        MS-SSIM \cite{wang2003multiscale}	&   0.563   &   0.486   &   0.501	&	0.462	\\
        VIF \cite{sheikh2006image}	        &   0.524   &   0.433   &   0.479	&	0.397	\\
        VSNR \cite{chandler2007vsnr}	    &   0.375   &   0.321   &   0.411	&	0.368	\\
        VSI \cite{zhang2014vsi}	            &   0.516   &   0.450   &   0.517	&	0.458	\\
        MAD \cite{larson2010most}	        &   0.626   &   0.608   &   0.580	&	0.543	\\
        NQM \cite{damera2000image}	        &   0.416   &   0.346   &   0.395	&	0.364	\\
        UQI \cite{wang2002universal}	    &   0.548   &   0.486   &   0.450	&	0.420	\\
        IFC \cite{sheikh2005information}	&   0.677   &   0.594   &   0.555	&	0.485	\\
        GSM \cite{liu2012image}	            &   0.469   &   0.418   &   0.465	&	0.409	\\
        RFSIM \cite{zhang2010rfsim}	        &   0.304   &   0.266   &   0.328	&	0.304	\\
        SRSIM \cite{zhang2012sr}	        &   0.654   &   0.566   &   0.636	&	0.573	\\
        FSIM \cite{zhang2011fsim}	        &   0.561   &   0.467   &   0.571	&	0.504	\\
        FSIMc \cite{zhang2011fsim}	        &   0.559   &   0.468   &   0.573	&	0.506	\\
        NIQE \cite{mittal2012making}	    &   0.102   &   0.064   &   0.132	& 	0.034	\\
        MA \cite{ma2017learning}	        &   0.203   &   0.201   &   0.147	&	0.140	\\
        PI \cite{blau20182018}	            &   0.166   &   0.169   &   0.145	&	0.104	\\ \hline
        LPIPS-Alex \cite{zhang2018unreasonable}	&	0.646	&	0.628	&	0.571	&	0.566	\\
        LPIPS-VGG \cite{zhang2018unreasonable}	&	0.647	&	0.591	&	0.633	&	0.595	\\
        PieAPP \cite{prashnani2018pieapp}	&	0.697	&	0.706	&	0.597	&	0.607	\\
        WaDIQaM \cite{bosse2017deep}	    &	0.654	&	0.678	&	0.548	&	0.553	\\
        DISTS \cite{ding2021comparison}	    &	0.686	&	0.674	&	0.687	&	0.655	\\
        SWD \cite{pipal}                    &	0.668	&	0.661	&	0.634   &   0.624   \\ \hline
        IQT (ours)                          &  $\textbf{0.741}$   &  $\textbf{0.718}$   &     -       &   -       \\
        IQT-C (ours)                         & $\textbf{0.876}$ & $\textbf{0.865}$ & $\textbf{0.790}$ & $\textbf{0.799}$ \\ \hline
	\end{tabularx}
\end{table}
\endgroup

\thispagestyle{empty}
\begin{figure}[b]
    \centering
	\includegraphics[width=.82\linewidth]{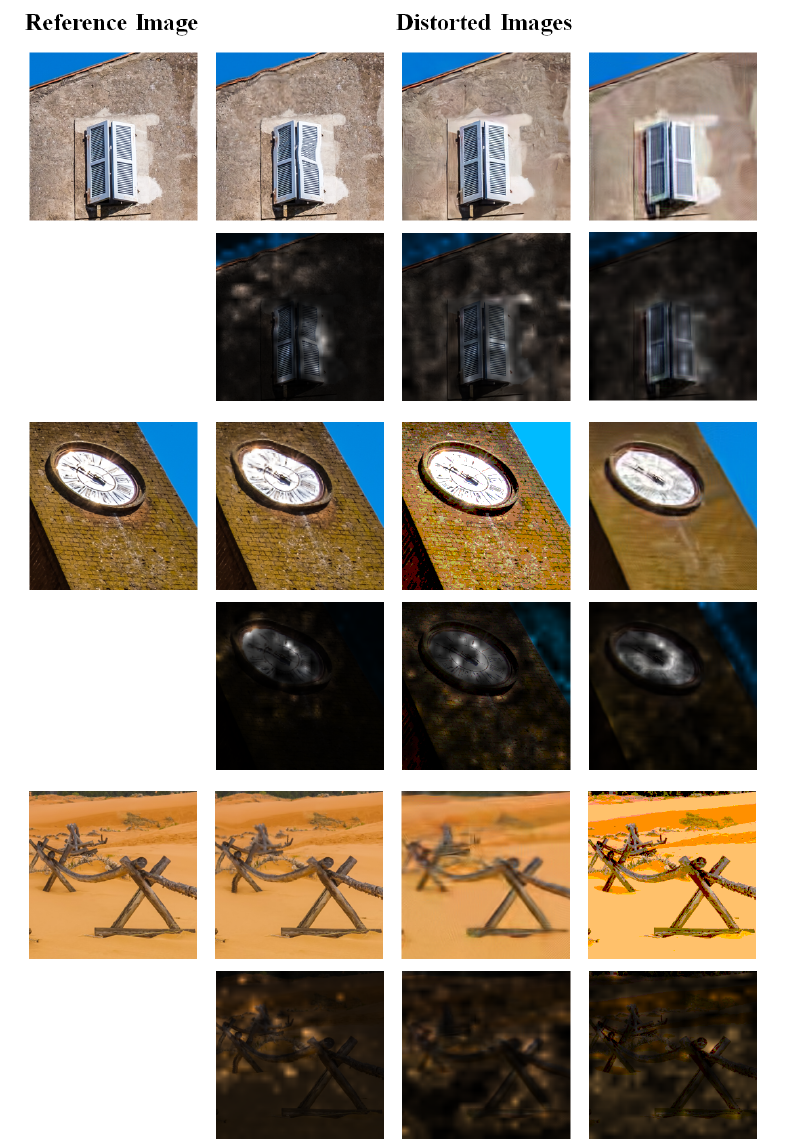}
	\caption{Visualization of attention maps from the proposed IQT. The center-cropped images are randomly sampled from the PIPAL \cite{pipal} dataset. Attention maps are averaged over all attention weights in the encoder and decoder.}
	\label{fig:attn}
\end{figure}
\thispagestyle{empty}

Our model is also evaluated on PIPAL \cite{pipal} dataset.
The IQT trained on KADID-10k dataset shows the best performance among all metrics.
The benchmark results comparing the existing IQA methods on PIPAL validation and testing datasets are shown in Table \ref{tb:bench4}.
Corresponding scatter plots of the predicted scores of IQT and the ground-truth MOS for validation dataset are also presented in Fig. \ref{fig:scatter1}.
It is shown that our method could also be a promising approach in the field of quality assessment on various datasets including generative models' output images.
Moreover, as shown in the results on the standard IQA datasets, our model shows a robust performance on different dataset.

\begingroup
\renewcommand{\arraystretch}{.92}
\begin{table*}[ht]
\small
\centering
\caption{\label{tb:abl1_input} Comparison of performance on three standard IQA databases depending on the inputs to the transformer encoder and decoder. The top performing method is highlighted in bold face.}
\begin{tabularx}{0.9\linewidth}{@{\extracolsep{\fill}}c|ccc|ccc|ccc}
	\hline
	\multirow{2}{*}{No.}&\multicolumn{3}{c|}{Encoder}&\multicolumn{3}{c|}{Decoder}&LIVE&CSIQ&TID2013 \\ \cline{2-10}
	&$F_{dist}$ & $F_{ref}$ & $F_{diff}$ & $F_{dist}$ & $F_{ref}$ & $F_{diff}$ & SRCC/KRCC&SRCC/KRCC&SRCC/KRCC \\ \hline
    (1)&\checkmark & & & \checkmark & &	& 0.901 / 0.713 & 0.768 / 0.575 & 0.646 / 0.468   \\ 
	(2)&\checkmark & & & & \checkmark &	& 0.934 / 0.767	& 0.855 / 0.670 & 0.739 / 0.548   \\ 
    (3)&& \checkmark & & \checkmark & &	& 0.954 / 0.805 & 0.865 / 0.680 & 0.755 / 0.564   \\ \hline
	(4)&\checkmark & & & & & \checkmark	& 0.967 / 0.838	& 0.944 / 0.803 & 0.884 / 0.698   \\ 
	(5)&& \checkmark & & & & \checkmark & 0.967 / 0.837	& 0.945 / 0.803 & 0.881 / 0.694   \\ \hline
	(6)&& & \checkmark & \checkmark & & & 0.969 / 0.843	& 0.945 / 0.803 & 0.897 / \textbf{0.714}   \\ 
	(7)&& & \checkmark & & \checkmark & & \textbf{0.970} / \textbf{0.845} & \textbf{0.947} / \textbf{0.805} & \textbf{0.896} / 0.712   \\ 
	(8)&& & \checkmark & & & \checkmark & 0.968 / 0.840	& 0.942 / 0.795 & 0.889 / 0.704   \\ \hline	
\end{tabularx}
\end{table*}
\endgroup

Fig. \ref{fig:attn} shows the examples of attention maps from the IQT model.
It refers to the area where the model focuses more when predicting the perceptual quality.
From our model architecture, the learned attention weights exist in the MHA layers of the encoder and decoder.
We visualize the attention maps by averaging all of the attention weights and resizing to the image size.
It is observed that the attention is spatially localized or spread uniformly across whole image depending on the image and distortion type.
It is important to see the entire image and, then, focus on a localized region when one perceives the quality of the image.
Our approach to determine the important region based on the self-attention mechanism will be useful to predict the quality.


\thispagestyle{empty}
\subsection{Ablations}

The use of the difference information between reference and distorted images is one of the important factors in the proposed architecture.
As mentioned in the previous section \ref{sec3}, the input into the encoder or decoder is a feature embedding and there are three types available, i.e., $F_{ref}$, $F_{dist}$, and $F_{diff}$.
To investigate the effect of input types and location, we conduct ablation experiment and the results of performance comparison are shown in Table \ref{tb:abl1_input}.

First, it is found that the use of the difference feature embedding as the input is a better choice than using only reference and distorted feature embeddings directly on the input.
It is shown that the models (4)-(8) have better performance than models (1)-(3) in Table \ref{tb:abl1_input}.
From this experiment, the model (7) is selected for our model design and this means that the $F_{diff}$ and $F_{ref}$ are used into the encoder and decoder, respectively.
When difference information enters the encoder or decoder, there is no significant performance difference between putting distorted or reference feature embedding in the other side.
We can find the similar results between the models (4) and (5), and between the models (6) and (7).
From this experiment, it is concluded that the difference information is an important factor in the proposed architecture for the IQA task.

\begingroup
\renewcommand{\arraystretch}{1}
\begin{table}[ht]
\small
\centering
\caption{\label{tb:abl2} Comparison of performance on the three standard IQA databases according to the method of using difference information. ``Feature'' refers to a difference operation conducted between feature maps extracted from the backbone. ``Image'' refers to the difference operation on RGB images.}
\begin{tabularx}{\linewidth}{@{\extracolsep{\fill}}cccc}
	\hline
	\multirow{2}{*}{Diff. Info.}&LIVE&CSIQ&TID2013\\ \cline{2-4}
	&SRCC/KRCC&SRCC/KRCC&SRCC/KRCC\\ \hline
	Feature	&	0.970 / 0.845 	&	0.947 / 0.805	&	0.896 / 0.712   \\
	Image	&	0.954 / 0.809	&	0.946 / 0.798   &	0.862 / 0.671   \\ \hline
\end{tabularx}
\end{table}
\endgroup

An additional experiment is conducted to prove that the difference information in the feature level is more effective than that in the image level.
The comparison results are shown in Table \ref{tb:abl2}.
Application of the difference information in the feature level that is important in our model design results in a better performance for all datasets.
In other words, the difference information in perceptual space is more useful to predict an image quality score compared to the RGB color space.

\thispagestyle{empty}
\subsection{NTIRE 2021 Perceptual IQA Challenge}

This work is proposed to participate in the NTIRE 2021 perceptual image quality assessment challenge \cite{gu2021ntire}.
The objective of this challenge is to develop a model predicting a value with high accuracy comparable to the ground-truth MOS.
The PIPAL \cite{pipal} dataset is used for the NTIRE 2021 challenge.
For this challenge, we train our model on training dataset provided in the NTIRE 2021 challenge.
The same model structure and both training and testing strategies are applied for the challenge.
The model hyper-parameters are set as follow:  $L=1$, $D=128$, $H=4$, $D_{feat}=1024$, and $D_{head}=128$.
The input image size of the model is set to $192\times192\times3$; therefore, we set the number of patches in feature map $N=441$.
The model for the NTIRE 2021 challenge is denoted as IQT-C to distinguish from the previously mentioned model IQT in Tables \ref{tb:bench1}, \ref{tb:bench4} and \ref{tb:bench3}.

\begingroup
\renewcommand{\arraystretch}{.85}
\begin{table}[ht]
    \small
	\centering
	\caption{\label{tb:bench3} Performance comparison of the participants on testing dataset of the NTIRE 2021 challenge. Main score is calculated as the sum of PLCC and SRCC. The number in the parenthesis denotes the rank. Only a few of the teams are shown in this Table. This result is provided from the NTIRE 2021 IQA challenge report \cite{gu2021ntire}.}
	\begin{tabularx}{0.45\textwidth}{@{\extracolsep{\fill}}llll}
		\hline
		Entries &    PLCC    &   SRCC    & Main Score $\uparrow$ \\ \hline
        IQT-C (ours) &	0.7896	(1)	&	0.7990	(2)	&	1.5885	(1)	\\
        Anonymous 1 &	0.7803	(2)	&	0.8009	(1)	&	1.5811	(2)	\\
        Anonymous 2 &	0.7707	(4)	&	0.7918	(3)	&	1.5625	(3)	\\
        Anonymous 3 &	0.7709	(3)	&	0.7770	(4)	&	1.5480	(4)	\\
        Anonymous 4 &	0.7615	(5)	&	0.7703	(6)	&	1.5317	(5)	\\
        Anonymous 5 &	0.7468	(7)	&	0.7744	(5)	&	1.5212	(6)	\\
        Anonymous 6 &	0.7480	(6)	&	0.7641	(7)	&	1.5121	(7)	\\ \hline
	\end{tabularx}
\end{table}
\endgroup

\begin{figure}[ht]
    \small
    \centering
	\includegraphics[width=0.6\linewidth]{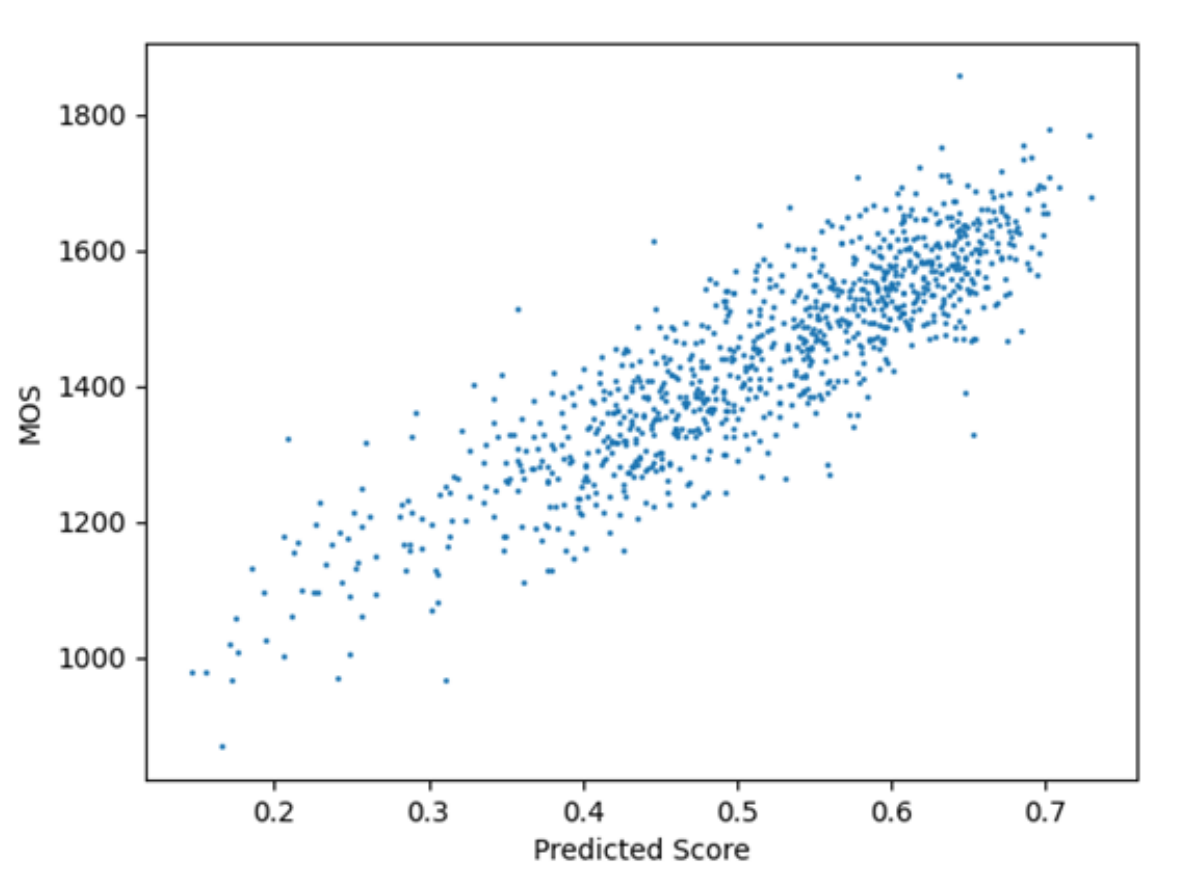}
	\caption{Scatter plots of ground-truth mean opinion scores (MOSs) against predicted scores of IQT-C on the PIPAL validation dataset.}
	\label{fig:scatter2}
\end{figure}

The benchmark results of the IQT-C on validation and testing datasets of the NTIRE 2021 challenge are shown in Table \ref{tb:bench4}.
The scatter plot is also illustrated in Fig. \ref{fig:scatter2}.
The IQT-C shows the best performance among all metrics.
In addition, a better performance than the IQT model trained on the KADID-10k is also found.
Final result of the challenge in testing phase is reported in Table \ref{tb:bench3}.
The rankings of the entries are determined in terms of main score, which is calculated with summation of PLCC and SRCC.
Our model won the first place in terms of the main score among all participants.
In terms of PLCC and SRCC, we obtain the first and second highest scores, respectively.

The model trained on the PIPAL shows outstanding performance for the validation and testing dataset in Table \ref{tb:bench4}.
However, on the other hands, it tends to increase the risk of over-fitting.
When we evaluate the IQT-C model on the three standard IQA datasets, it shows much lower performance than the IQT trained on KADID-10k (Table \ref{tb:bench1}).
It is noteworthy noting that the IQT-C is the special case of our approach for the NTIRE 2021 challenge.
However, there is a room for improvement in terms of robustness for any other distortion types when we train the IQT on PIPAL dataset.
In addition, future work is needed to improve the model to solve this problem.

\thispagestyle{empty}
\section{Conclusion}
\label{sec5}
We proposed an IQT and it is appropriately applied to a perceptual image quality assessment task by taking an advantage of transformer encoder-decoder architecture.
The IQT demonstrated the outstanding performance on the three standard IQA databases compared to many existing methods.
Our method also showed the best performance for the latest IQA dataset that contains deep learning-based distorted images.
The IQT showed another promising example that the transformer based approach can achieve a high performance even in the perceptual quality assessment task.

Despite the success of our model, there exists a room for improvement.
Further investigation of the transformer based approach, especially considering more diverse resolutions and distortion types, is needed.
In addition, developing a no-reference metric for perceptual quality assessment will be desirable that can be used in real-world scenarios.

\thispagestyle{empty}
\renewcommand{\thepage}{}
{\small
\bibliographystyle{ieee_fullname}
\bibliography{cvpr_mrc}
}
\thispagestyle{empty}
\end{document}